\documentclass[manuscript]{acmart}

\AtBeginDocument{%
  \providecommand\BibTeX{{%
    \normalfont B\kern-0.5em{\scshape i\kern-0.25em b}\kern-0.8em\TeX}}}

\setcopyright{acmcopyright}
\copyrightyear{2021}
\acmYear{2021}
\acmDOI{10.1145/3466171}

\acmJournal{TALLIP}
\acmVolume{21}
\acmNumber{1}
\acmArticle{14}
\acmMonth{11}

\usepackage{hyperref}
\usepackage{xcolor}
\usepackage{rotating}
\usepackage{arabtex}
\usepackage{utf8}
\setcode{utf8}

\begin{document}

\title{Empirical evaluation of shallow and deep learning classifiers for Arabic sentiment analysis}


\author{Ali~Bou~Nassif}
\affiliation{%
  \institution{Computer Engineering Department, University of Sharjah}
  \city{Sharjah}
  \country{UAE}}
\email{anassif@sharjah.ac.ae}

\author{Abdollah~Masoud~Darya}
\affiliation{%
  \institution{Electrical Engineering Department, University of Sharjah}
  \city{Sharjah}
  \country{UAE}}
\email{abdollah.masoud@ieee.org}

\author{Ashraf~Elnagar}
\affiliation{%
  \institution{Computer Science Department, University of Sharjah}
  \city{Sharjah}
  \country{UAE}}
\email{ashraf@sharjah.ac.ae}

\renewcommand{\shortauthors}{Nassif, et al.}

\begin{abstract}
This work presents a detailed comparison of the performance of deep learning models such as convolutional neural networks (CNN), long short-term memory (LSTM), gated recurrent units (GRU), their hybrids, and a selection of shallow learning classifiers for sentiment analysis of Arabic reviews. Additionally, the comparison includes state-of-the-art models such as the transformer architecture and the araBERT pre-trained model. The datasets used in this study are multi-dialect Arabic hotel and book review datasets, which are some of the largest publicly available datasets for Arabic reviews. Results showed deep learning outperforming shallow learning for binary and multi-label classification, in contrast with the results of similar work reported in the literature. This discrepancy in outcome was caused by dataset size as we found it to be proportional to the performance of deep learning models. The performance of deep and shallow learning techniques was analyzed in terms of accuracy and F1 score. The best performing shallow learning technique was Random Forest followed by Decision Tree, and AdaBoost. The deep learning models performed similarly using a default embedding layer, while the transformer model performed best when augmented with araBERT.
\end{abstract}



\maketitle

\section{Introduction}
\label{intro}
With the rise of Web 2.0 and the ubiquity of online social networks, massive amounts of opinion-based data are generated by users that list their opinions and personal views online. Analyzing such data would produce valuable information regarding key trends, product evaluations, stock market predictions, and public opinion surveys. Sentiment analysis (SA) is a field that aims to extract sentiments relayed in pieces of text based on its contents. Unfortunately, automated systems of SA have faced challenges in accurately labeling human sentiment due to the complex nature of the semantics used in human language. Moreover, the Arabic language adds another hurdle to automated SA due to its large number of dialects, its morphological richness and its ingrained ambiguity \cite{al2015deep,elnagar2021systematic}.\par
Arabic-speaking internet users have grown by  9,348\%\footnote{\url{https://www.internetworldstats.com/stats7.htm}} in the last twenty years, marking the largest growth value in any language. Furthermore, following English, Mandarin and Spanish, Arabic speakers are the fourth largest online linguistic population, tallied at 5.2\% of the total online population\footnote{\url{https://www.internetworldstats.com/stats7.htm}}. Despite being one of the fastest-growing languages in terms of online users, the field of Arabic SA is still not as mature as its English counterpart. However, the last few years have shown growing interests in the field of Arabic SA \cite{al2019comprehensive}. In addition to the task of SA \cite{alnawas2019sentiment}, also referred to as opinion mining \cite{badaro2014large,baly2017characterization}, there is also great interest in the fields of emotion mining \cite{badaro2018ema,mohammad2018semeval} and text classification \cite{elnagar2020arabic}.\par
This work conducts a comprehensive comparison of some of the most widely used deep and shallow learning (SL) techniques using two of the largest publicly available datasets for multi-dialectal Arabic reviews: Hotel Arabic Reviews Dataset (HARD) \cite{elnagar2018hotel} and the Book Reviews in Arabic Dataset (BRAD) \cite{Elnagar2016,elnagar2018annotated} (see section 3.1).  It compares four of the main deep learning (DL) models, which requires further study in the field of Arabic natural language processing \cite{al2018deep2,Lulu2018deep,almuzaini2020impact}.\par
Section \ref{LR} presents the main contributions found in the literature of Arabic SA using DL classifiers. In section \ref{Metho}, the methodology and datasets used in this work are discussed, as well as the pre-processing steps followed. Section \ref{R&D} discusses the results obtained, along with the learning curve. Finally, the conclusion summarizes the findings of this study and proposes potential avenues for future work.\par

\subsection{Research Objectives}
The results of this work prove that DL models are superior to SL models for SA of Arabic reviews, given that the dataset used is of adequate size. This is in contrast to the work done in \cite{al2018deep1}, which found that support vector machines (SVM) performed significantly better than convolutional neural networks (CNN) and recurrent neural networks (RNN) for the task of Arabic sentiment polarity identification.  We hypothesize that the findings of \cite{al2018deep1} were a result of using a dataset of an inadequate size that does not utilize the full potential of DL techniques, and therefore that they deliver an unfair comparison. Not only does this work consider binary sentiment classification, it also compares between the best performing SL and DL classifiers for multi-label classification (5-label), which strengthens the comparison established between SL and DL methods.\par
The contributions of this work are summarized as follows:
\begin{itemize}
  \item To the best of our knowledge, this is the first work that comprehensively compares the performance of SL and DL techniques for the purposes of Arabic SA. Not only is the evaluation done for the three main DL architectures (CNN, LSTM and GRU) but also all possible combinations of the aforementioned techniques. Furthermore, the state-of-the-art transformer architecture was studied and compared against the previously mentioned DL models.
  \item In this paper we prove that when a dataset of adequate size is used, DL classifiers outperform SL classifiers.
   \item We establish the point at which the best DL technique outperforms the best performing SL technique by plotting the learning curve for various dataset sizes. This was done for both binary and 5-label sentiment classification.
   \item We conduct a study highlighting the effect of different embedding layers on the performance of the transformer model. The comparison includes default embeddings derived from the analyzed dataset,  custom embeddings acquired from Arabic websites, as well as pre-trained models such as araBERT.
  \item This is the first study to comprehensively evaluate both the HARD and BRAD datasets, which are some of the largest publicly available Arabic review datasets.
\end{itemize}

\section{Literature Review}
\label{LR}
Arabic SA is a highly challenging research area that involves complex tasks. The tasks that have been the most frequent subjects of study are subjectivity classification, sentiment classification, lexicon creation, aspects extraction, aspect sentiment classification and opinion spam detection  \cite{badaro2015light,boudad2017sentiment,hamdi2018clasenti,beseiso2020subword,badaro2020link,elnagar2016investigation}.\par
The work done in the field of Arabic SA is vast and encompasses many fields and niche applications which cannot be covered in this section, thus the reader is referred to the following surveys that present the most recent updates in the field   \cite{al2019comprehensive,al2018deep2,dohaiha2018deep,badaro2019survey,nassif2020deep}.  However, with that being said, there are few contributions that will be highlighted in the following paragraph that are relevant to the scope of this work.\par
The authors in \cite{al2015deep} investigated the results of using DL for sentence-level Arabic SA. In their work, they used Deep Belief Networks (DBN), Deep Autoencoders (DAE), Deep Neural Networks (DNN) and Recursive Auto Encoder (RAE).  They used an annotated collection of 1,180 sentences selected from the Data Consortium Arabic Tree Bank (LDCATB) dataset. Their approach showed that RAE performed best out of the four models. The authors then worked on improving the results of the RAE model and introduced AROMA \cite{al2017aroma}: a recursive DL model for opinion mining in Arabic. An improvement in classification accuracy by upto 12\% was achieved by performing word sentiment embedding and generating syntactic parse trees to be used as a reference to AROMA's recursion. The authors in \cite{baly2017sentiment} used Recursive Neural Tensor Networks (RNTN) for SA of Arabic text. They  used a subset of the Qatar Arabic Language Bank (QALB) data set, consisting of 1,177 sentences. They claim that the proposed approach outperforms SVM, RAE, and LSTM. In \cite{al2017hybrid}, the authors use CNN and LSTM networks with word embedding features on the Arabic Sentiment Tweets Dataset (ASTD), which consists of 10,000 sentences. In their evaluation, they found that the LSTM performed best, with an accuracy of 87\%. In \cite{baly2017omam}, the authors compared an English state-of-the-art method to classify binary sentiment in Arabic tweets against a cluster-based sentiment classification approach and RAE. Their results show that the English state-of-the-art method outperformed the competition. A common trend seen in most related work is their reliance on datasets composed of fewer than 1000 sentences, while DL techniques demand the use of vast and comprehensive datasets.\par
The first use of hybrid DL techniques was done recently in the work presented in \cite{alayba2018combined} where a combined CNN and LSTM architecture was used to analyze binary sentiment in several Twitter datasets. The largest dataset used in this study contains 2,500 sentences, which resulted in accuracy in the mid-90-s for tweet-level SA.\par

\section{Methodology}
\label{Metho}
In this section we start by introducing the selected datasets. We then discuss the pre-processing steps taken to prepare the data for the SA task, and we describe the SL and DL techniques used in our empirical evaluation.
\subsection{Datasets}
\paragraph{HARD} The first dataset used is the Hotel Arabic Reviews Dataset (HARD) \cite{elnagar2018hotel}, which contains 373,750 reviews\footnote{compiled from \url{booking.com}} (see Table \ref{HARD} for a sample of the reviews). Two variants of the HARD dataset are available: balanced and unbalanced. The unbalanced dataset was chosen as the target of this study,  because it is the larger of the two. The number of unique tokens in the HARD dataset is in the order of 90K. Note: the variables contained in the HARD dataset are listed in Table \ref{table1}. What makes this dataset unique is that it combines reviews in Modern  Standard Arabic (MSA) and Dialectal Arabic (DA). It is also the largest publicly available dataset of Arabic hotel reviews.\par

\begin{table}[]
\caption{Two sample reviews from HARD with their literal English translation.} \label{HARD}
\begin{tabular}{|c|l|}
\hline
Rating & \multicolumn{1}{c|}{Review} \\ \hline
4 & \begin{tabular}[c]{@{}l@{}}\<استثنائي. سعر جدا مناسب وأتمنى ان لا يزيد فلمستقبل حتى نظل عملائكم ونظافة الغرف وخدمة البوفيه>\\ Exceptional. Price very reasonable and I wish it does not increase in the future so that we remain your\\customers and cleanliness of rooms and buffet service\end{tabular} \\ \hline
2 & \begin{tabular}[c]{@{}l@{}}\<كانت سيئه جدا بسبب سوء المعامله. الجو. كانت تجربه فاشله لاني تعرضت لسوء تعامل من موظغي الاستقبال>\\ It was very bad due to the bad treatment. The weather. It was a failed experiment because I was\\mistreated by the reception staff.\end{tabular} \\ \hline
\end{tabular}
\end{table}

\begin{table}[]
\caption{HARD dataset variables.} \label{table1}
\begin{tabular}{|l|l|l|}
\hline
Column \# & Variable     & More Information                    \\ \hline
1         & Rating       & Possible values are 1-5             \\ \hline
2         & Review ID    & Unique ID assigned to each review   \\ \hline
3         & Hotel ID     & ID assigned to each hotel           \\ \hline
4         & User ID      & Unique ID assigned to each user     \\ \hline
5         & Room ID      & Unique ID assigned to each room     \\ \hline
6         & \# of Nights & Number of nights                    \\ \hline
7         & Review       & Reviewers opinion written in MSA/DA \\ \hline
\end{tabular}
\end{table}

\paragraph{BRAD} The Book Reviews in Arabic Dataset (BRAD) \cite{elnagar2018annotated} consists of a collection containing 508,538 book reviews\footnote{taken from \url{www.goodreads.com}} in total (see Table \ref{BRAD} for a sample of the reviews). Similar to HARD, BRAD contains reviews in MSA and DA. The BRAD dataset is made up of around 220K unique tokens. Note: the variables contained in the BRAD dataset are listed in Table \ref{table2}. These datasets are considered to be two of the largest available for the applications of Arabic SA and machine learning.\par

\begin{table}[]
\caption{Two sample reviews from BRAD with their literal English translation.} \label{BRAD}
\begin{tabular}{|c|l|}
\hline
Rating & \multicolumn{1}{c|}{Review} \\ \hline
2 & \begin{tabular}[c]{@{}l@{}}\<الروايه مفيهاش احداث كتير . ممله لحد كبير. بس الاسلوب في الكتابه كان حلو جدا.>\\ \< بس كنت متوقعه الروايه احلي من كده>\\ The novel does not have many events. Boring to a large extent. But the style of writing is very good.\\ But I was expecting the novel better than this.\end{tabular} \\ \hline
5 & \begin{tabular}[c]{@{}l@{}}\<احمد مراد بأسلوبه المبدع يتألق فى سرد ورواية التاريخ و يلقى الضوء على شخصيات فى ظل>\\ \< هذه الحقبه التاريخيه والأحداث>\\ Ahmad Murad (the author) with his creative style shines in narrating history and casting a light on\\ characters under this historical era and events.\end{tabular} \\ \hline
\end{tabular}
\end{table}

\begin{table}[]
\caption{BRAD dataset variables.} \label{table2}
\begin{tabular}{|l|l|l|}
\hline
Column \# & Variable  & More Information                    \\ \hline
1         & Rating    & Possible values are 1-5             \\ \hline
2         & Review ID & Unique ID assigned to each review   \\ \hline
3         & Book ID   & ID assigned to each book            \\ \hline
4         & User ID   & Unique ID assigned to each user     \\ \hline
5         & Review    & Reviewers opinion written in MSA/DA \\ \hline
\end{tabular}
\end{table}

\subsection{Pre-processing} 
The following steps were taken to clean the data and prepare it for further processing. Note: Every step applies to both datasets unless otherwise specified.\par
\begin{enumerate}
\item [Step.1] The datasets were checked for missing values/nulls, none were found. 
\item [Step.2] All variables other than rating and review were removed from the datasets.
\item [Step.3] For binary classification, in the rating column, values corresponding to $\{1,2\}$ were changed to 0 and values corresponding to $\{4,5\}$ were changed to 1. Values corresponding to 3 were dropped from the dataset. The problem thus was transformed into a binary classification problem, with the outcome being either positive (1) or negative (0). For 5-label classification, all rating values from $1-5$ have been included.
\item [Step.4] Symbols, numerals and various other characters were removed from the text. While brackets \<( )> on Arabic and English keyboards look similar, they are represented by different characters. This also applies to periods \<(.)> and commas \<(,)>. Both versions of each character were removed.
\item [Step.5] Normalization was performed automatically by scanning the text and replacing some characters with others so that words that may have been spelled differently would now be spelled the same. Example: the Arabic word for ``something" \<(شيء)> could be incorrectly represented as, \<(شي, شئ, شىء)> therefore, the ending characters \<(يء, ىء, ئ)> had to be made similar, i.e., converted to \<(ي)>. Normalization will have varying effects on different dialects.
\item [Step.6] Outliers were removed. Two checks were set up to remove outliers. For the first check, a sentence could not exceed 500 characters or be less than 3 characters. The value of 500 was chosen as it was found that 98\% of sentences were less than that value, while the other 2\% reached 1,500 characters. Having more characters will negatively impact the classification process \cite{li2018textual}. The minimum value 3 was chosen because some of the sentences had single word reviews such as the Arabic word for ``good" \<(جيد)>, these were important for classification. For the second check, a sentence could not have more than 100 words or less than 1 word. These values were found to be appropriate, for the reasons discussed above and as recommended in \cite{li2018textual}.
\item [Step.7] Characters that were repeated more than twice in a single word were reduced to two repetitions. For example, the exaggerated Arabic word for ``wonderful" \<(راااااااائع)> was reduced to \<(راائع)>. Another check was set to change \<(اا, وو, يي)> that were repeated twice to be single characters, as it was found that these characters were the most often repeated. This step prevents the deletion of repeated characters that are meant to be repeated, such as the \<(م)> in the Arabic word for ``excellent" \<(ممتاز)>.
\item [Step.8] The dataset was then balanced to ensure that our testing parameters were not biased. For the binary classification of HARD, since the initial dataset was unbalanced, with  46,001 of the reviews being negative and 252,252 being positive, a random sample of 45,500 was taken from both sides. Later, both were combined to form a balanced dataset of 91,000 reviews. As BRAD was a larger dataset,  60,000 samples were taken randomly from 63,104 negative and 249,737 positive reviews to form a balanced dataset of 120,000 reviews to be used for binary classification. For 5-label classification, the HARD dataset was balanced around the smallest label, which was the lowest rating of 1. Since rating 1 contained around 12,500 reviews, random samples of the same size (12,500) were taken from each label, adding up to a full dataset of 62,500 reviews. Note that this process was implemented after all the previous data cleaning and pre-processing steps.  The training size was selected to be 90\% of the balanced dataset, while 10\% was selected for testing using the holdout technique. Out of the 90\% training set, 10\% was selected for validation for DL, and 9-fold cross-validation for SL. Furthermore, the performance metrics of the DL techniques were obtained as an average of 10 runs, where the dataset is randomly resampled at the beginning of each run.
\item [Step.9]  Finally, tokenization was implemented, by splitting the text into separate tokens (words). Next, the integer vector was defined for each token and each sentence was padded to a length of 100 tokens.
\end{enumerate}
Note: no other preprocessing techniques were used as they could bias the results towards a specific model and therefore provide an unfair advantage, i.e., feature selection and extraction \cite{tubishat2019improved,khan2019enswf}. Furthermore, several techniques generally seen in preliminary trial and error tests to either reduce performance or leave it unchanged at the cost of added computation\textemdash such as stemming \cite{almuzaini2020impact} or removing stopwords\textemdash were not implemented. The resultant datasets were then tested and compared using SL and DL techniques.\par

\subsection{Shallow Learning}

\textbf{Decision Tree (DT)}: Decision tree \cite{safavian1991survey} is a classification algorithm that divides a set of data into smaller sets based on tests that are defined at each node in the tree. The tree consists of a root node, internal nodes and terminal nodes (leaves). Each node in the tree has one parent node and two or more child nodes. Based on this, each data set is classified by assigning it a location according to the framework defined by the tree.\par
\textbf{Random Forest (RF)}: A random forest \cite{breiman2001random} consists of a combination of tree predictors where each tree is dependent on the variables of a random vector sampled independently, with all the trees in the forest having the same distribution. In random forests, the best split is used among a randomly chosen subset of predictors at each node. This adds robustness against overfitting and improved performance at classification \cite{liaw2002classification}.\par
\textbf{Support Vector Machines (SVM)}: Support Vector Machines (SVM) \cite{breiman2001random} is the most widely used machine learning network for two-group classification problems. Conceptually, input vectors are mapped non-linearly to a feature space with high dimensionality. A linear decision surface plane or line is constructed in the feature space to separate the two classes also known as a hyperplane. In this work, SVM was implemented using two kernels: linear (lin.) and radial basis function (RBF).\par
\textbf{k-nearest neighbors (KNN)}: $K$-Nearest Neighbours (KNN) is a simple classification technique \cite{altman1992introduction}. For instance, to classify a data record $ntest$, $K$ nearest neighbors are saved to form the neighborhood for $ntest$. The classification is based on the most frequent nearest neighbors, either with or without distance weighting. But to apply $K$NN an appropriate value of $K$ needs to be considered which makes the technique biased.\par
\textbf{Multilayer Perceptron (MLP)}: Multi-Layer Perceptron consists of one input layer of input neurons followed by one or more hidden layers and a single output layer \cite{rosenblatt1960perceptrons}. Each layer is composed of nodes or neurons that are fully connected to nodes in the subsequent layers up to the output layers. The activation function for the initial input layer is linear with no thresholds. Subsequently, however, the hidden node layers have non-linearity in their activation function in addition to the threshold. The output layer activity is dependent on linear function and threshold. Each hidden unit node is related to the addition of the weighted sum of every input node in the initial layer and the associated threshold. The output layer is associated with the hidden layers in the same manner.\par
\textbf{Gaussian Na\"ive Bayes (GNB)}: The Na\"ive Bayes (NB) classifier applies the Bayes theorem with the assumption that each pair of features are independent. The NB classifier finds the probability that a given instance belongs to a certain class. Gaussian Na\"ive Bayes, on the other hand, executes the classification by assuming the likelihood of the features to be Gaussian.\par
\textbf{AdaBoost (AB)}: The Adaptive Boosting (AdaBoost) algorithm \cite{freund1997decision} is one of the first proposed boosting algorithms and works by combining several relatively weak and inaccurate models to create an accurate prediction model. AdaBoost can be used to substantially reduce learning algorithms errors and is mainly aimed at classification applications \cite{freund1996experiments}.\par
\subsection{Deep Learning}
\textbf{CNN}: Convolutional Neural Networks (CNN) \cite{lecun1988theoretical} are feedforward neural networks originally conceived for the field of computer vision and have shown to be effective for natural language processing (NLP) applications \cite{kim2014convolutional}. They utilize a layer with convolving filters applied to local features. They feature convolution in place of general matrix multiplication, which is present in standard neural networks. This decreases the number of weights, thereby reducing the complexity of the network, causing it to be one of the best-performing DL techniques in terms of execution time. Furthermore, another advantage of CNN is that it requires minimal preprocessing. Along with its low complexity, this paved the way for its use in NLP, speech and handwriting recognition, and image classification, amongst many others \cite{liu2017survey}.\par
\textbf{LSTM}: Long short term memory (LSTM) \cite{hochreiter1996bridging} network is a type of recurrent neural networks (RNN) that is effective at learning problems related to sequential data. It tackles these problems by capturing long-term temporal dependencies. LSTM does not suffer from the optimization issues affecting the basic form of RNN due to its complex nature and the repetition of its modules \cite{schmidhuber2015deep}. The basic idea behind the LSTM architecture is a memory cell that maintains its state over time and nonlinear gating units that control information flow in and out of the cell \cite{greff2016lstm}. It also features three main gates: the input, forget and output gates. The input block is connected to the output block and all the gates.\par
\textbf{GRU}: The gated recurrent unit (GRU) framework was proposed by \cite{cho2014properties} in 2014. Similar to LSTM, GRU contains gating units that control the flow of information. While in LSTM networks, the gate controls the amount of memory that is utilized by other units in the network, in GRU all contents are exposed without any restriction. It has been reported, however, that GRU outperforms LSTM for nearly all tasks except language modeling \cite{jozefowicz2015empirical}. Furthermore, the gap between the performance of LSTM and GRU networks can be minimized by initializing LSTM’s forget gate bias to one.  GRU was previously used in several Arabic NLP tasks such as \cite{al2018emojis}.\par
\begin{figure*}
\fbox{\includegraphics[width=0.65\textwidth]{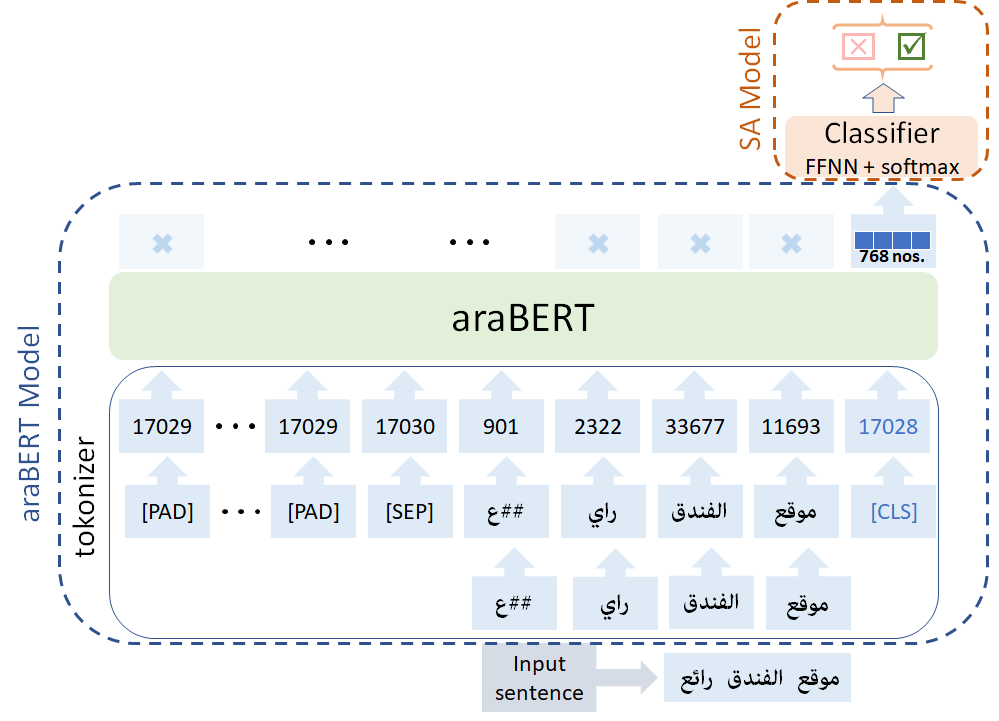}}
\caption{The workflow is summarized in the pipeline diagram above showing how the sentiment classifier (the fine-tuning task) is added on top of the araBERT transformer model.}
\label{fig:trans}
\end{figure*}

 
\textbf{Transformer}: The transformer (TRANS) model was first proposed in \cite{vaswani2017attention}. The transformer consists of an encoder and a decoder. The input sequence is taken by the encoder and mapped into a higher dimensional space. The decoder then produces an output sequence from the mapped input. It has been reported to train significantly faster than recurrent and convolutional architectures for translation tasks \cite{vaswani2017attention}. Transformers (feed-forward architecture) allow for efficient training on huge datasets, with the objective of simply predicting words based on their context. Building such models is very expensive; however, a variety of models are published and ready to use in downstream tasks such as SA. It has been reported that fine-tuning these models on some smaller, supervised datasets can improve classification results. This process is called transfer learning. Therefore, instead of building a transformer model from scratch, we will simply take an existing one (in this case: araBERT \cite{baly2020arabert}) and utilize its parameters to initialize the sentiment classifier and achieve the result (see Figure \ref{fig:trans}).\par 

This work evaluates the three main DL architectures discussed above. Those are CNN, LSTM and GRU, and all of their possible combinations,  as well as the TRANS model. Note that bidirectional LSTM (Bi-LSTM) has been used in place of LSTM, as it performed better in preliminary trial and error tests. Training, testing and pre-processing were executed on Google Colab (\url{https://colab.research.google.com/}), with a Tesla K80 GPU, an Intel Xeon processor and 12 GB of RAM.\par

\section{Results and Discussion}
\label{R&D}
In this study, the performance of the techniques discussed above was tested using some of the most commonly used performance metrics. These are accuracy, precision, recall, F1, area under the curve of the receiver operating characteristics, training time and testing time \cite{davis2006relationship}. Accuracy is simply the number of correct predictions divided by the total number of predictions. Accuracy can be represented as follows:
\begin{equation}
Accuracy=\frac{TP + TN}{TP + TN + FP + FN}
\end{equation}
where the sum of true positives and true negatives is divided by the sum of true positives, true negatives, false positives and false negatives. True positives (TP) are positive examples labeled correctly as positives. True negatives (TN) are negative examples labeled correctly as negative. False positives (FP) are negative examples labeled incorrectly as positive. False negatives (FN) are positive examples labeled incorrectly as negative \cite{davis2006relationship}.
Precision is defined as:
\begin{equation}
Precision=\frac{TP}{TP + FP}
\end{equation}
and recall can be presented as:
\begin{equation}
Recall=\frac{TP}{TP + FN}
\end{equation}
The F1 measure can be deduced from the following:
\begin{equation}
F1=2\times\frac{Precision\times Recall}{Precision + Recall}
\end{equation}
The area under the curve (AUC) of the receiver operating characteristics (ROC) presents an important metric for binary classification, as it portrays the ability of the system to distinguish between good and bad reviews \cite{davis2006relationship}. The best-performing system would have AUC values closer to one, while the worst-performing systems would have AUC values closer to zero.  The parameters of all models used are presented in section \ref{appen}. For multi-label classification, other metrics are used such as the macro/micro/weighted average for precision, recall and F1 \cite{madjarov2012extensive}. We note, however, that since we are using a balanced dataset, the metrics obtained from the macro/micro/weighted average would yield similar results. Therefore, for simplification, we are referring to the precision, recall and F1 for multi-label classification as average precision, average recall and average F1 respectively.\par

\subsection{Shallow Learning}\label{SL}
\subsubsection{Binary Classification}
\begin{figure*}
\fbox{\includegraphics[width=0.85\textwidth]{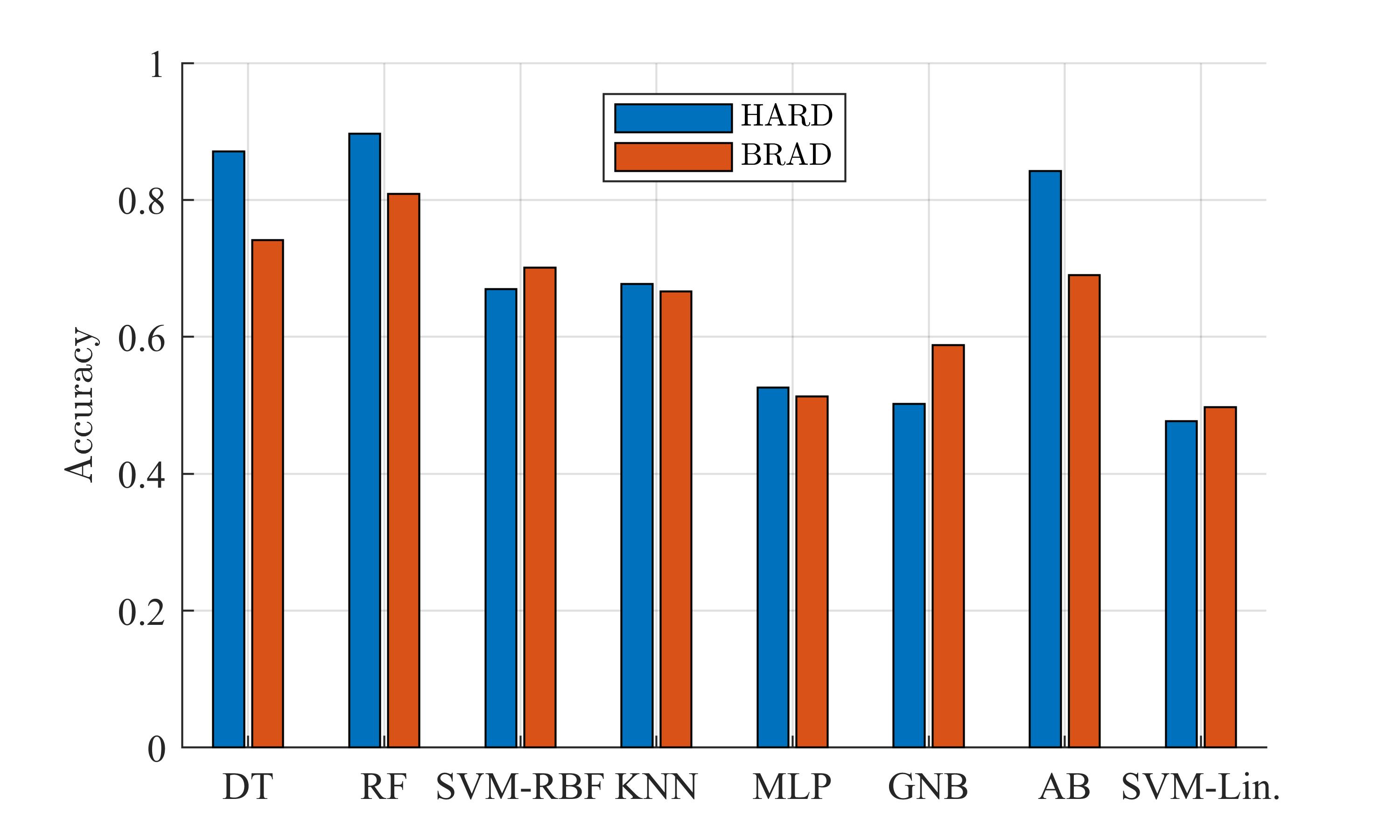}}
\caption{Comparison between the binary classification accuracy obtained for SL techniques for both datasets.}
\label{fig:6}
\end{figure*}

The results in Table \ref{table3} and Figure \ref{fig:6} clearly show the superiority of three SL techniques from the rest. Random Forest performed the best, followed by Decision Tree and AdaBoost. It can also be seen that in most cases, the SL techniques performed best when being trained and tested on the HARD dataset as compared to BRAD, even though BRAD is a larger dataset than HARD (120K vs 90K reviews). This can be attributed to the size of the vocabulary being utilized in each dataset. Where BRAD contains unique vocabulary in the order of 220K, HARD contains only 90K, less than half that amount. This agrees with the findings of \cite{li2018textual}, where it was found that  datasets with shorter reviews and less unique vocabulary performed best.  We note that the RBF kernel SVM outperformed the linear kernel SVM. This means that the data analyzed was not linearly separable. We also highlight that other studies that utilize significantly smaller datasets found SVM to be the best performing SL classifier \cite{baly2017sentiment,shoukry2012sentence}. Using our larger datasets we prove that ensemble learning methods such as RF, DT and AB perform best.\par

\begin{table}[]

\caption{Binary classification results from training and testing the HARD (H) and BRAD (B) datasets using SL. Note: all results are rounded to three significant digits. \textbf{Bold} results represent the best performance from each category.} \label{table3}
\begin{tabular}{|l|c|c|c|c|c|c|c|c|c|c|}
\hline
 & \multicolumn{2}{c|}{Accuracy} & \multicolumn{2}{c|}{Precision} & \multicolumn{2}{c|}{Recall} & \multicolumn{2}{c|}{F1} & \multicolumn{2}{c|}{AUC} \\ \hline
Classifier & H & B & H & B & H & B & H & B & H & B \\ \hline
DT & 0.871 & 0.741 & 0.868 & 0.732 & 0.876 & \textbf{0.747} & 0.872 & 0.740 & 0.870 & 0.740 \\ \hline
RF & \textbf{0.897} & \textbf{0.809} & \textbf{0.891} & \textbf{0.887} & 0.905 & 0.700 & \textbf{0.898} & \textbf{0.783} & \textbf{0.900} & \textbf{0.810} \\ \hline
SVM (RBF) & 0.670 & 0.701 & 0.607 & 0.675 & 0.858 & 0.431 & 0.711 & 0.526 & 0.670 & 0.700 \\ \hline
KNN & 0.677 & 0.666 & 0.680 & 0.680 & 0.672 & 0.605 & 0.676 & 0.640 & 0.680 & 0.660 \\ \hline
MLP & 0.526 & 0.513 & 0.611 & 0.667 & 0.149 & 0.018 & 0.240 & 0.036 & 0.530 & 0.500 \\ \hline
GNB & 0.502 & 0.588 & 0.501 & 0.673 & \textbf{0.980} & 0.315 & 0.663 & 0.429 & 0.500 & 0.580 \\ \hline
AB & 0.842 & 0.690 & 0.823 & 0.676 & 0.872 & 0.712 & 0.846 & 0.693 & 0.840 & 0.690 \\ \hline
SVM (Linear) & 0.477 & 0.497 & 0.481 & 0.493 & 0.573 & 0.779 & 0.523 & 0.603 & 0.480 & 0.500 \\ \hline
\end{tabular}
\end{table}

\subsubsection{5-label Classification}
\begin{figure*}
\fbox{\includegraphics[width=0.85\textwidth]{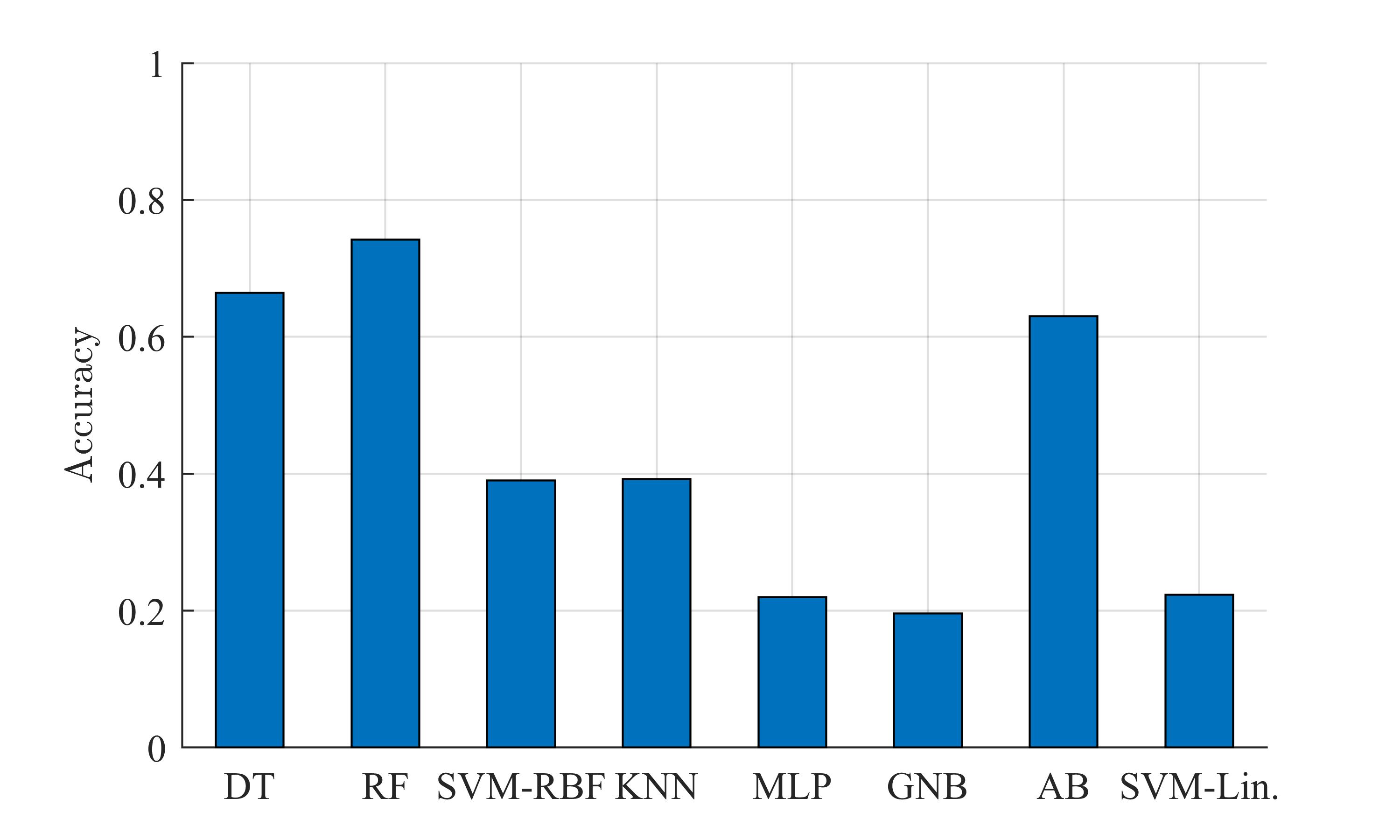}}
\caption{Comparison between the 5-label classification accuracy obtained for shallow learning techniques for the HARD dataset.}
\label{fig:MCSL}
\end{figure*}

Since the SL classifiers performed better when trained and tested on the HARD dataset for binary sentiment classification, the next step was to evaluate the classifiers for 5-label sentiment classification using the same dataset. The results in Table \ref{tableMC_SL} and Figure \ref{fig:MCSL} show the performance of SL classifiers on the task of 5-label sentiment classification. Similar to the binary classification case, Random Forest performed best, followed by Decision Tree and AdaBoost. This further shows the superiority of these ensemble methods. The other classifiers performed considerably worse, with the GNB classifier performing worse than the baseline accuracy of 0.2.

\begin{table}[]

\caption{5-label classification results from training and testing the HARD dataset using SL. Note: all results are rounded to three significant digits. \textbf{Bold} results represent the best performance from each category.} \label{tableMC_SL}
\begin{tabular}{|l|c|c|c|c|c|}
\hline
Classifier & Accuracy & Avg. Precision & Avg. Recall & Avg. F1 & Avg. AUC \\ \hline
DT & 0.664 & 0.660 & 0.660 & 0.660 & 0.790 \\ \hline
RF & \textbf{0.742} & \textbf{0.740} & \textbf{0.740} & \textbf{0.740} & \textbf{0.839} \\ \hline
SVM (RBF) & 0.390 & 0.500 & 0.390 & 0.370 & 0.617 \\ \hline
KNN & 0.392 & 0.390 & 0.390 & 0.390 & 0.619 \\ \hline
MLP & 0.220 & 0.200 & 0.210 & 0.140 & 0.509 \\ \hline
GNB & 0.196 & 0.240 & 0.200 & 0.080 & 0.500 \\ \hline
AB & 0.630 & 0.660 & 0.630 & 0.630 & 0.768 \\ \hline
SVM (Linear) & 0.223 & 0.230 & 0.220 & 0.200 & 0.513 \\ \hline
\end{tabular}
\end{table}

\subsection{Deep Learning}\label{DL}
\subsubsection{Binary Classification}
\begin{figure*}
\fbox{\includegraphics[width=0.85\textwidth]{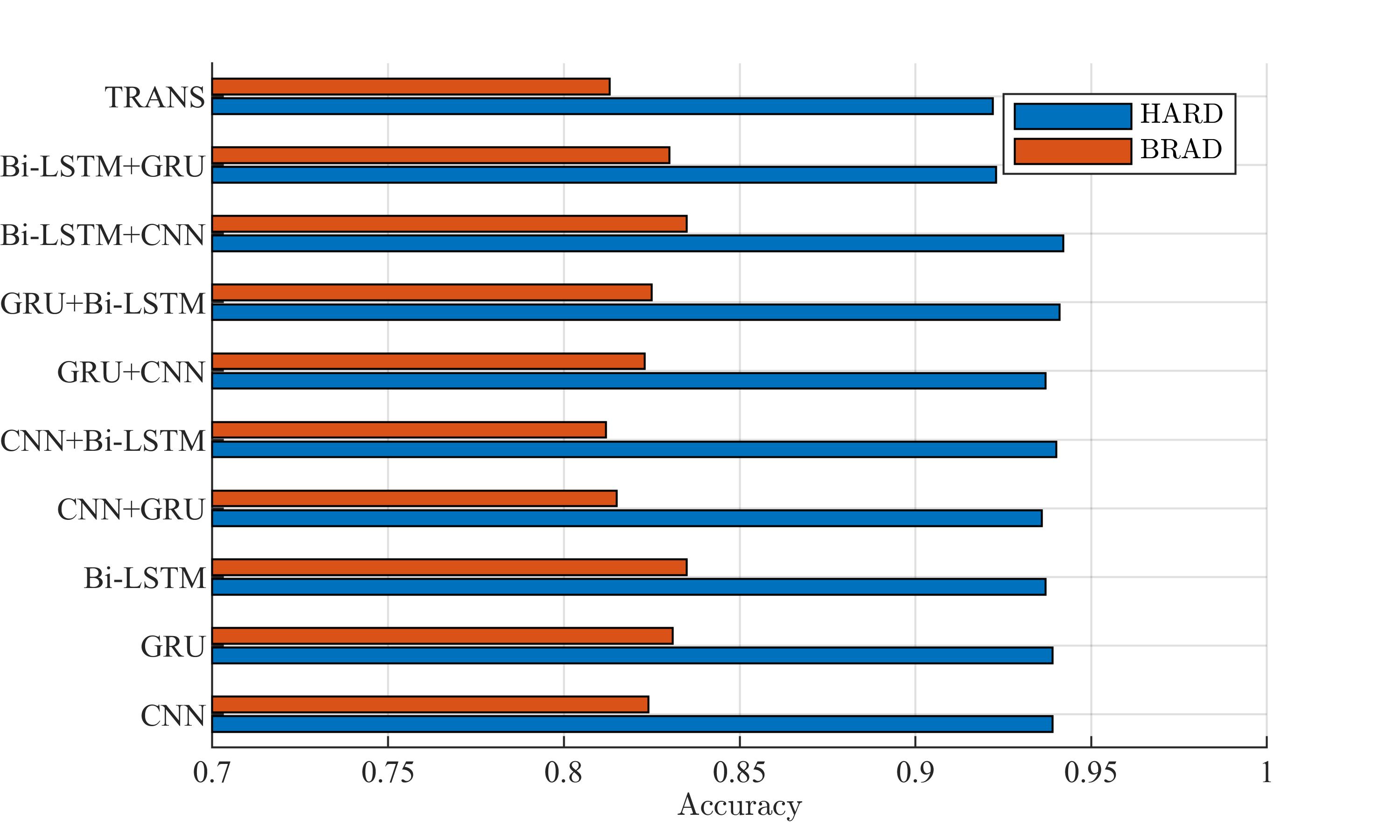}}
\caption{Comparison between the binary classification accuracy obtained for deep learning techniques for both datasets.}
\label{fig:7}
\end{figure*}

The previously mentioned DL techniques were tested in a similar manner to the SL classifiers (see Table \ref{table4}). Accuracy is seen to be around the mid-90-s for the HARD dataset and around 82\% for the BRAD dataset. The results are fairly close for all techniques (see Figure \ref{fig:7}).  The best-performing classifier in terms of accuracy was the Bi-LSTM+CNN hybrid model for both the HARD and BRAD datasets.\par
To check if the most accurate model is statistically different from the other tested models, we conducted the non-parametric Wilcoxon test \cite{gehan1965generalized} between the classifier with the highest accuracy value (the Bi-LSTM+CNN hybrid model) and other tested classifiers. Based on the $p$-values reported in Table \ref{table4}, we noticed that the Bi-LSTM+CNN classifier is statistically different, i.e., having a $p$-value that is lower than $0.05$, from 6 models based on a $95\%$ confidence interval. We also observed that the Bi-LSTM+CNN classifier is statistically different from the CNN classifier at a $90\%$ confidence interval. Thus we conclude that the Bi-LSTM+CNN hybrid classifier is statistically different from the majority of the tested DL classifiers.

\begin{table}[]
\caption{Binary classification results from training and testing the HARD (H) and BRAD (B) datasets using DL. Note: all results are rounded to three significant digits. \textbf{Bold} results represent the best performance from each category.} \label{table4}
\begin{tabular}{|l|c|c|c|c|c|c|c|c|c|c|c|}
\hline
 & \multicolumn{2}{c|}{Accuracy} & \multicolumn{2}{c|}{Precision} & \multicolumn{2}{c|}{Recall} & \multicolumn{2}{c|}{F1} & \multicolumn{2}{c|}{AUC} & \multicolumn{1}{c|}{$p$-value}\\ \hline
Classifier & H & B & H & B & H & B & H & B & H & B & H\\ \hline
CNN & 0.939 & 0.824 & 0.933 & 0.820 & 0.947 & 0.824 & 0.939 & 0.822 & 0.939 & 0.824 & 0.0883 \\ \hline
GRU & 0.939 & 0.831 & 0.924 & \textbf{0.827} & \textbf{0.956} & 0.831 & 0.940 & 0.829 & 0.939 & 0.831 & 0.0342 \\ \hline
Bi-LSTM & 0.937 & \textbf{0.835} & 0.931 & 0.825 & 0.945 & \textbf{0.844} & 0.938 & \textbf{0.834} & 0.939 & \textbf{0.835} & 0.0008 \\ \hline
CNN+GRU & 0.936 & 0.815 & \textbf{0.934} & 0.804 & 0.939 & 0.826 & 0.937 & 0.815 & 0.937 & 0.814 & 0.0211 \\ \hline
CNN+Bi-LSTM & 0.940 & 0.812 & 0.933 & 0.804 & 0.949 & 0.816 & 0.941 & 0.810 & 0.940 & 0.813 & 0.2120 \\ \hline
GRU+CNN & 0.937 & 0.823 & 0.928 & 0.819 & 0.948 & 0.823 & 0.938 & 0.820 & 0.937 & 0.822 & 0.0045 \\ \hline
GRU+Bi-LSTM & 0.941 & 0.825 & 0.932 & 0.815 & 0.952 & 0.836 & 0.942 & 0.825 & \textbf{0.941} & 0.826 & 0.7623 \\ \hline
Bi-LSTM+CNN & \textbf{0.942} & \textbf{0.835} & 0.931 & 0.826 & 0.955 & 0.842 & \textbf{0.943} & \textbf{0.834} & \textbf{0.941} & 0.834 & - \\ \hline
Bi-LSTM+GRU & 0.923 & 0.830 & 0.904 & 0.821 & 0.949 & 0.838 & 0.926 & 0.829 & 0.924 & 0.830 & 0.0003 \\ \hline
TRANS & 0.922 & 0.813 & 0.918 & 0.801 & 0.928 & 0.825 & 0.923 & 0.813 & 0.923 & 0.813 & \textbf{0.0002} \\ \hline
\end{tabular}
\end{table}

\subsubsection{5-label Classification}
\begin{figure*}
\fbox{\includegraphics[width=0.85\textwidth]{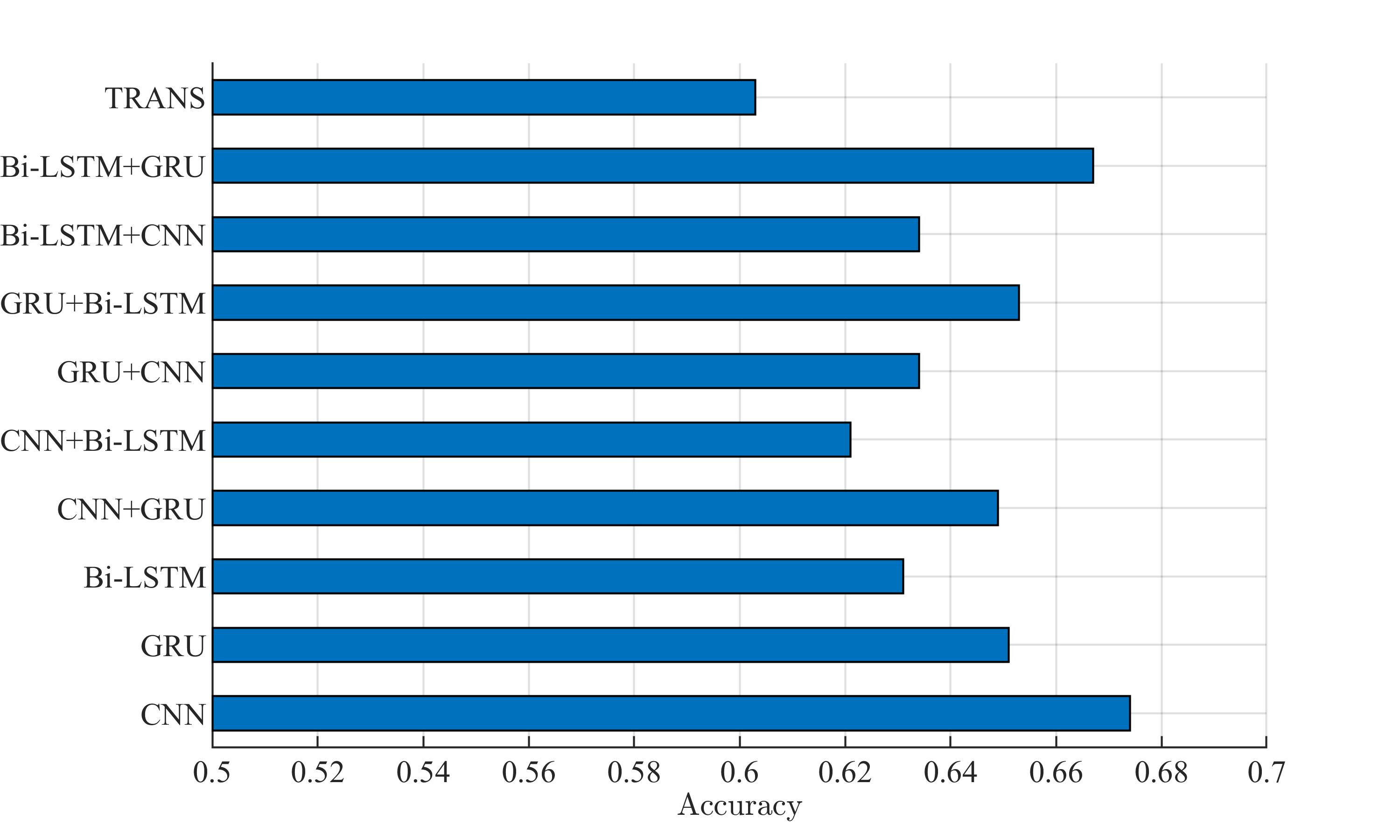}}
\caption{Comparison of classification accuracy in 5-label sentiment analysis obtained for deep learning techniques for the HARD dataset.}
\label{fig:MCDL}
\end{figure*}

As the DL classifiers performed better when trained and tested on the HARD dataset for binary sentiment classification, the next step was to evaluate the classifiers for 5-label sentiment classification using the same dataset. The results in Table \ref{tableMC_DL} and Figure \ref{fig:MCDL} show the performance of DL classifiers on the task of 5-label sentiment classification. We find that the DL classifiers have similar performance as they did in the binary classification task, with classification accuracy being measured at around 0.65 and with CNN having the highest accuracy at 0.674. However, we note that while the DL classifiers outperform most of their SL counterparts for the task of 5-label classification, the RF classifier achieved higher accuracy at 0.742. This point will be discussed in greater detail in the following section.

\begin{table}[]

\caption{5-label classification results from training and testing the HARD dataset using deep learning. Note: all results are rounded to three significant digits. \textbf{Bold} results represent the best performance from each category.} \label{tableMC_DL}
\begin{tabular}{|l|c|c|c|c|c|}
\hline
Classifier & Accuracy & Avg. Precision & Avg. Recall & Avg. F1 & Avg. AUC \\ \hline
CNN & \textbf{0.674} & \textbf{0.773} & \textbf{0.674} & \textbf{0.720} & \textbf{0.812} \\ \hline
GRU & 0.651 & 0.754 & 0.653 & 0.698 & 0.798 \\ \hline
Bi-LSTM & 0.631 & 0.725 & 0.632 & 0.674 & 0.785 \\ \hline
CNN+GRU & 0.649 & 0.754 & 0.648 & 0.697 & 0.797 \\ \hline
CNN+Bi-LSTM & 0.621 & 0.715 & 0.621 & 0.664 & 0.779 \\ \hline
GRU+CNN & 0.634 & 0.770 & 0.633 & 0.697 & 0.793 \\ \hline
GRU+Bi-LSTM & 0.653 & 0.713 & 0.655 & 0.681 & 0.793 \\ \hline
Bi-LSTM+CNN & 0.634 & 0.727 & 0.635 & 0.678 & 0.787 \\ \hline
Bi-LSTM+GRU & 0.667 & 0.709 & 0.667 & 0.687 & 0.799 \\ \hline
TRANS & 0.603 & 0.683 & 0.602 & 0.640 & 0.766 \\ \hline
\end{tabular}
\end{table}

\subsection{Learning Curve}
\subsubsection{Binary Classification}
An experiment was conducted to understand the effect of dataset size on the accuracy of the best-performing DL and SL models, i.e., Bi-LSTM+CNN and RF. This was done by training and testing the best-performing techniques using the HARD dataset at varying dataset sizes (see Table \ref{table5}). The changes in accuracy can then be captured and compared. To ensure the true behavior of the techniques is observed, 100 random samples were extracted from the dataset and split by (80-10-10) for training, testing and validation (note that cross-validation was used for  RF). The models were trained using the split data and the mean accuracy and standard deviation values for the 100 runs were recorded in Table \ref{table5}.\par

\begin{table}[]
\caption{The performance of Bi-LSTM+CNN and RF versus dataset size variation for binary classification of HARD. Accuracy is averaged over 100 runs, where each run randomly resamples the dataset.} \label{table5}
\begin{tabular}{|l|l|c|c|c|c|}
\hline
\multicolumn{1}{|c|}{\begin{tabular}[c]{@{}c@{}}Dataset\\ Size\end{tabular}} & \multicolumn{1}{c|}{\begin{tabular}[c]{@{}c@{}}Number of\\ Reviews\end{tabular}} & \multicolumn{2}{c|}{Bi-LSTM+CNN} & \multicolumn{2}{c|}{RF} \\ \hline
\multicolumn{2}{|c|}{} & \begin{tabular}[c]{@{}c@{}}Mean\\ Accuracy\end{tabular} & \begin{tabular}[c]{@{}c@{}}Standard\\ Deviation\end{tabular} & \begin{tabular}[c]{@{}c@{}}Mean\\ Accuracy\end{tabular} & \multicolumn{1}{l|}{\begin{tabular}[c]{@{}l@{}}Standard\\ Deviation\end{tabular}} \\ \hline
1\% & 910 & 0.895 & 0.0327 & 0.921 & 0.0248 \\ \hline
5\% & 4,550 & 0.915 & 0.0140 & 0.919 & 0.0121 \\ \hline
10\% & 9,100 & 0.926 & 0.0095 & 0.913 & 0.0104 \\ \hline
50\% & 45,500 & 0.938 & 0.0038 & 0.902 & 0.0046 \\ \hline
100\% & 91,000 & 0.942 & 0.0026 & 0.899 & 0.0035 \\ \hline
\end{tabular}
\end{table}

\begin{figure*}
\fbox{\includegraphics[width=0.75\textwidth]{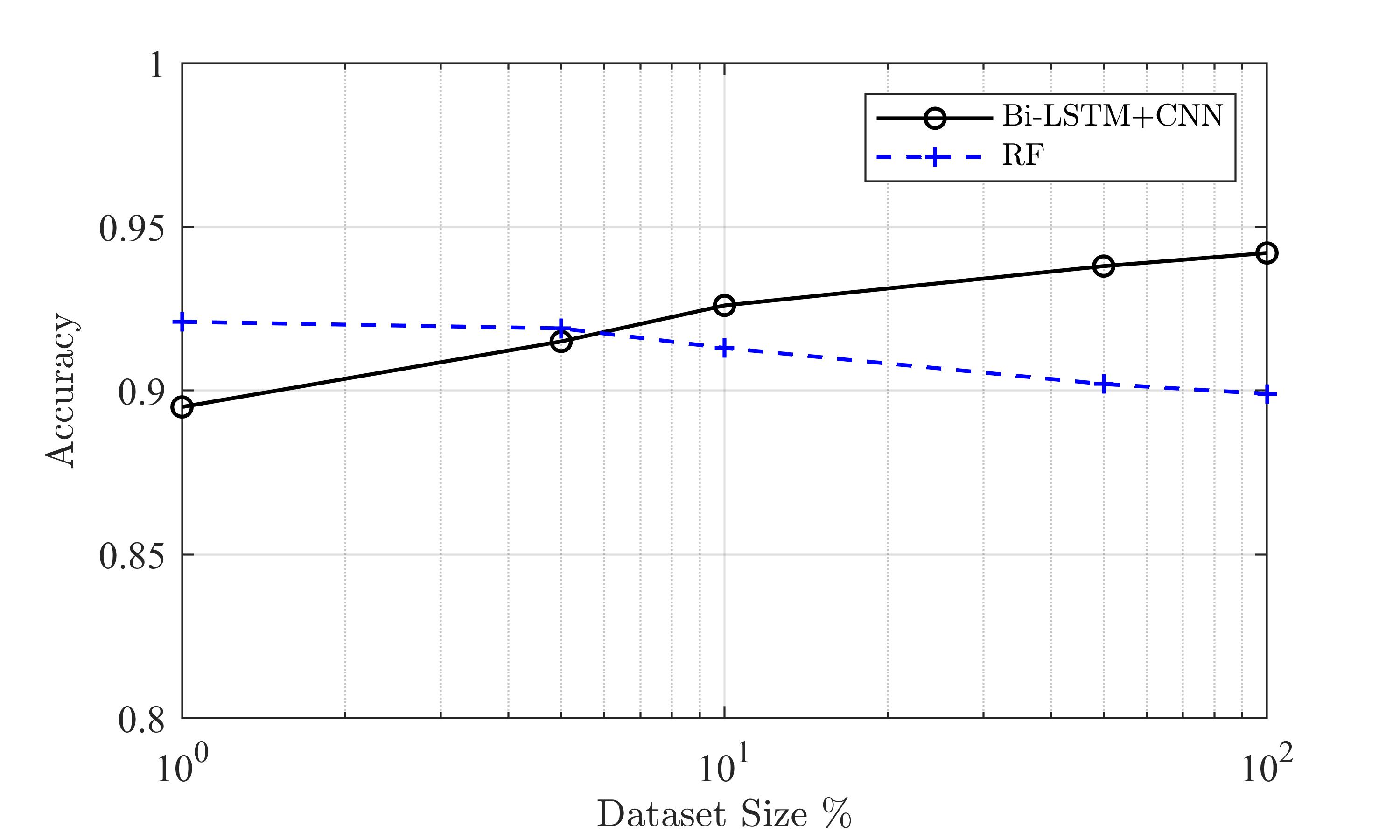}}
\caption{Binary classification accuracy versus the percentage of the dataset used for Bi-LSTM+CNN and RF. Accuracy is averaged over 100 runs, where each run randomly resamples the dataset. Note that the x-axis is in the logarithmic scale.}
\label{fig:8}
\end{figure*}

From Figure \ref{fig:8}, it can be seen that the mean accuracy of the models used is affected by dataset size variation.  As the utilized dataset size increased, so did the mean accuracy of the DL models. On the other hand, the SL classifier saw a gradual decrease in mean accuracy with dataset size increase. Furthermore, the standard deviation value for both classifiers is seen to decrease as the number of reviews increases. An important feature to note in this case is the percentage of the dataset needed for the mean accuracy of Bi-LSTM+CNN (a DL technique) to overtake the mean accuracy of RF (a SL technique), which is around 6\% of the dataset, corresponding to 5,460 reviews. Note that this value is highly dependent on the quality of the dataset used. The aforementioned feature is important in the argument for the use of DL techniques instead of SL methods for improved accuracy \cite{ng2015data}.\par

\subsubsection{Multi-label Classification}
A similar experiment was conducted for the case of multi-label classification instead of binary classification. This was also done using the HARD dataset, but instead of classifying positive/negative reviews (see Pre-processing, Step 3) we included all rating values from 1 to 5. Thus the problem became a 5-label classification problem. Similar to the previous experiment, a comparison was made between the best-performing deep and SL models, based on the results of Sections \ref{SL} and \ref{DL}; these were the CNN and RF classifiers. To ensure a fair comparison between the various labels, the dataset was balanced around the smallest label, which was the lowest rating of 1. Since rating 1 contained around 12,500 reviews, random samples of the same size (12,500) were taken from each label, adding up to a full dataset of 62,500 reviews. The data was then split (80-10-10) for training, testing and validation (note that cross-validation was used for  RF). Finally the mean accuracy and standard deviation values were obtained from 100 runs (where each run randomly samples the dataset before it is split) and recorded in Table \ref{table6}.\par

\begin{table}[]
\caption{The performance of CNN and RF versus dataset size variation for 5-label classification of HARD. Accuracy is averaged over 100 runs, where each run randomly resamples the dataset.} \label{table6}
\begin{tabular}{|l|l|c|c|c|c|}
\hline
\multicolumn{1}{|c|}{\begin{tabular}[c]{@{}c@{}}Dataset\\ Size\end{tabular}} & \multicolumn{1}{c|}{\begin{tabular}[c]{@{}c@{}}Number of\\ Reviews\end{tabular}} & \multicolumn{2}{c|}{CNN} & \multicolumn{2}{c|}{RF} \\ \hline
\multicolumn{2}{|c|}{} & \begin{tabular}[c]{@{}c@{}}Mean\\ Accuracy\end{tabular} & \begin{tabular}[c]{@{}c@{}}Standard\\ Deviation\end{tabular} & \begin{tabular}[c]{@{}c@{}}Mean\\ Accuracy\end{tabular} & \multicolumn{1}{l|}{\begin{tabular}[c]{@{}l@{}}Standard\\ Deviation\end{tabular}} \\ \hline
1\% & 625 & 0.481 & 0.1113 & 0.748 & 0.0538 \\ \hline
5\% & 3,125 & 0.587 & 0.0457 & 0.777 & 0.0238 \\ \hline
10\% & 6,250 & 0.615 & 0.0309 & 0.775 & 0.0194 \\ \hline
50\% & 31,250 & 0.656 & 0.0152 & 0.750 & 0.0074 \\ \hline
100\% & 62,500 & 0.668 & 0.0120 & 0.743 & 0.0060 \\ \hline
\end{tabular}
\end{table}

\begin{figure*}
\fbox{\includegraphics[width=0.75\textwidth]{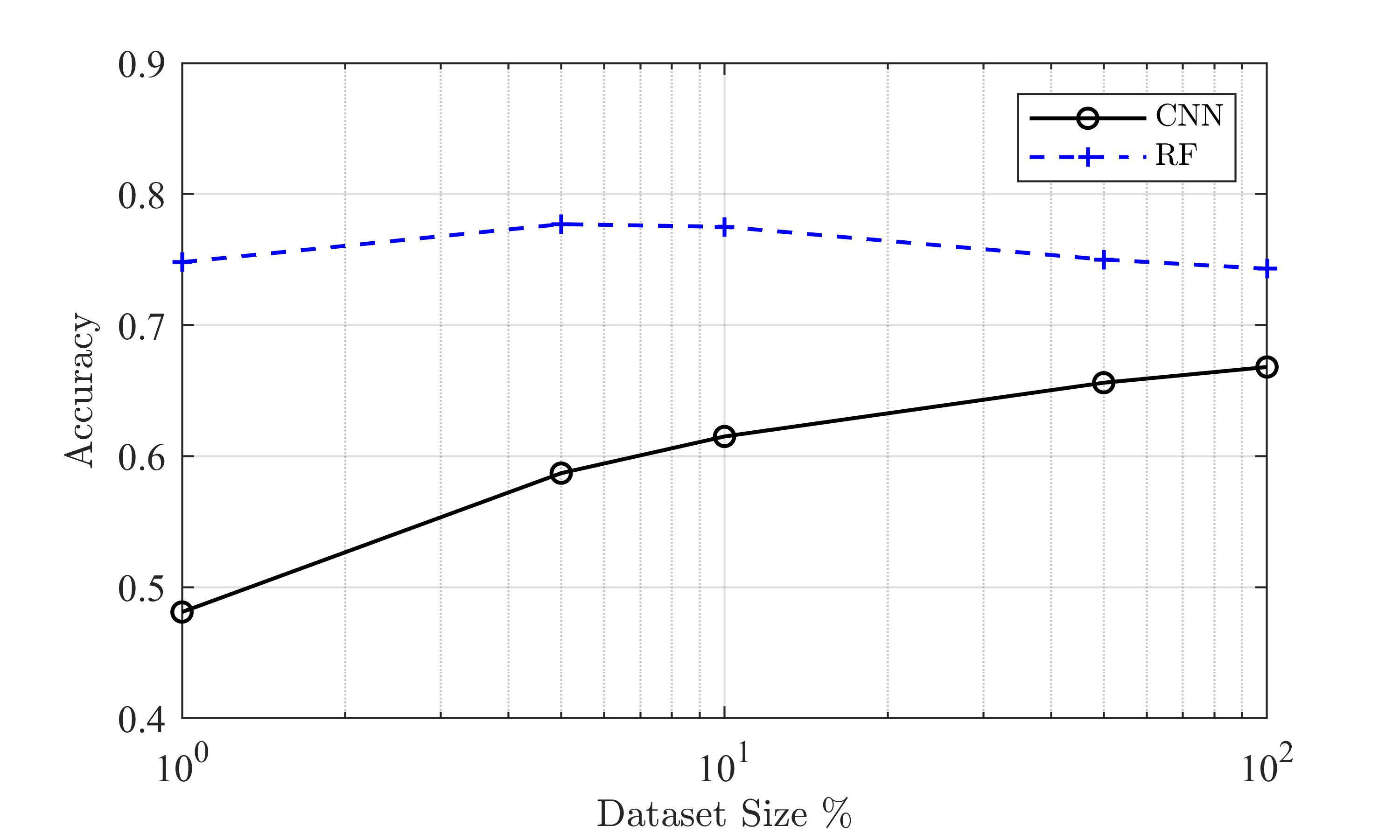}}
\caption{5-label classification accuracy versus the percentage of the dataset used for CNN and RF. Accuracy is averaged over 100 runs, where each run randomly resamples the dataset. Note that the x-axis is in the logarithmic scale.}
\label{fig:9}
\end{figure*}

Figure \ref{fig:9} presents the mean accuracy trend for both CNN and RF for a varying dataset size. We can see that the RF classifier outperforms the CNN classifier for all dataset sizes. Furthermore, similar to Figure \ref{fig:8}, it can be seen that as dataset size increases, so does the accuracy of CNN, whereas the performance of the RF classifiers remains consistent. Therefore, the CNN classifier will benefit from a larger dataset, allowing it to overtake the RF classifier in terms of classification accuracy.\par

\subsection{Embeddings}
 In the previous sections we have seen the TRANS model underperforming when compared to other DL architectures such as the Bi-LSTM+CNN hybrid model for binary classification and the CNN model for 5-label classification. This does not match recent findings in the literature \cite{eljundi2019hulmona}. We believe this is due to the fact that default single-layer token embeddings were used (vector size 64), whereas the TRANS model benefits greatly from more complex embedding layers.\par 
One of the main differences between SL and DL for NLP is the presence of an embedding layer in the front end of DL techniques and the lack thereof in SL techniques. For word embeddings, several options could be used, either learning word embeddings from the problem or reusing embeddings that are publicly available. Learning word embeddings from the problem can be done using three main methods. A model can learn the embeddings separately through a specific framework such as Word2Vec \cite{mikolov2013efficient},  fastText \cite{grave2018learning}, GloVe \cite{pennington2014glove} or the more recent araBERT \cite{baly2020arabert}. Alternatively, it can learn embedding jointly using an initial embedding layer or it can use sentiment specific embeddings \cite{altowayan2017improving}. We chose to use the second technique in this work (learning embedding jointly using an initial embedding layer) and compare it with state-of-the-art araBERT model (v0.1). The authors in \cite{baly2020arabert} obtained araBERT as a result of pre-training BERT for the Arabic language. They also found that it outperforms multilingual BERT (mBERT) \cite{devlin2018bert} for SA performed on the HARD dataset. We are also comparing the default embedding layers with a custom embedding layer, utilizing trigram multi-dialect embeddings taken from several websites \cite{einea2019sanad,elnagar2020arabic}. We used the TRANS architecture to compare the following:
\begin{itemize}
  \item Using a single default embedding layer for tokens
  \item Using a single custom embedding layer for tokens
  \item Using two default embedding layers, one for tokens and another for token positions
  \item Using a custom embedding layer for tokens and a default embedding layer for token positions
  \item Using araBERT
\end{itemize}

\begin{table}[]

\caption{The effect of embedding testing on 3K sentences for binary classification and 5K sentences for 5-label classification. The accuracy recorded is the mean accuracy measured over 100 re-runs, where each re-run randomly resamples the dataset.} \label{tableE1}
\begin{tabular}{|l|c|c|}
\hline
Embedding Type & Binary Accuracy & 5-Label Accuracy \\ \hline
Single Layer Default Token & 0.892 & 0.609 \\ \hline
Single Layer Custom Token & 0.894 & 0.616 \\ \hline
Two Layer Default Token + Position & 0.916 & 0.638 \\ \hline
Two Layer Custom Token + Position & 0.914 & 0.650 \\ \hline
araBERT & \textbf{0.927} & \textbf{0.736} \\ \hline
\end{tabular}

\end{table}

The results can be seen in Table \ref{tableE1} for binary and 5-label SA using the HARD dataset. We note that the chosen embedding vector size was 300. The results show the superiority of the araBERT model for binary and multi-label classification. Furthermore, we note the fact that using a two-layer token and position embedding had a greater improvement in performance than using a single-token embedding layer.  We also note that the custom embedding setup outperformed its default counterpart for multi-label classification. This could mean that the custom token embedding layer exploits label-specific phrases more effectively.
\par

\subsection{Misclassifications}

\begin{table}[]

\caption{Examples of misclassifications by the CNN model (binary classification).} \label{tableMs}
\begin{tabular}{|l|l|l|l|}
\hline
\# & Review                                                                                                                                                                                                                                                                                                                                                                                                                                                                                                                                                                                                                                                                                                                                                                                                                                      & Rating & \begin{tabular}[c]{@{}l@{}}Model \\ Prediction\end{tabular} \\ \hline
1 & \begin{tabular}[c]{@{}l@{}}
\<رحله قصيره الواي فاي سي قوه الما في دورات المياه ضعيف جدا>\\ 
Short trip bad Wi-Fi power of water in washroom weak\end{tabular}                                                                                                                                                                                                                                                                                                                                                                                                                                                                                                                                                                                                                                                                           & 1      & 0                                                           \\ \hline
2 & \begin{tabular}[c]{@{}l@{}}
\<استثناي كل شي ممتاز وراقي من جميع النواحي عدا الواي فاي>\\  
\<كلش ضعيف الواي فاي ضعيف جدا ولازم تعيد ربط الاتصال حتي اذا تنقلت>\\  
\<من غرفه الي اخري في نفس الشقه وكذلك لازم تعيد الاتصال ويطلب>\\
\<كلمه السر في كل مره تغادر وترجع الشقه>\\ 
Exceptional everything is excellent and elegant in all aspects except for the Wi-Fi\\ 
extremely weak Wi-Fi is very weak and you must reconnect even if you move \\ 
between rooms in the same apartment and you must reconnect and it asks for \\ 
password every time you leave and return to the apartment\end{tabular}                                                                                                                                                                                                                            & 1      & 0                                                           \\ \hline
3 & \begin{tabular}[c]{@{}l@{}}
\<جبد ولكن نبحث عن الافضل الموقع الخدمات لم تكن بالمستوي المطلوب مثل>\\ 
\<توفير مسلتزمات دورات المياه ايضا ارجو الاهتمام بالخدمات الفندقيه المقدمه وشكرا>\\ 
Good but we are looking for the best the location the services were \\ 
not up to par such as providing washroom amenities also please pay \\ 
attention to the provided hotel services and thank you\end{tabular}                                                                                                                                                                                                                                                                                                                                                                                                                         & 0      & 1                                                           \\ \hline
4 & \begin{tabular}[c]{@{}l@{}}
\<الكويت مكان ممتاز عايلي يمتاز في للهدو والخدمه الممتازه والنظافه>\\ 
\<وخدمه الضيافه القهوه العربيه متازه تاخير في استلام الغرفه الجيك ان>\\ 
Kuwait is an excellent place familial excels in quietness \\ 
and excellent service and cleanliness and hospitality Arabic\\ 
coffee excellent late in receiving the room check in\end{tabular}                                                                                                                                                                                                                                                                                                                                                                                                                                                          & 0      & 1                                                           \\ \hline
5 & \begin{tabular}[c]{@{}l@{}}
\<ارجو منهم ان يحافظوا علي نظافه الفندق والاهتمام بالخدمه الفطار>\\ 
\<متنوع والخدمه داخل بوفيه الفطار ممتازين لقد قمت بالحجر مبكرا وعندما قدمت قال لي>\\ 
\<ناسف لا يمكننا ان نوفر لك سرير مزدوج فاعطاني سريرين منفصلين معنا هذه المره الثانيه>\\ 
\<لي اسكن في هذا الفندق فالخدمه بالمره الاولي كانت افضل بسبب عدم الزحمه بالفندق>\\ 
I implore them to keep the hotel clean and pay attention to the service \\ 
breakfast was diverse and service at the breakfast buffet excellent I made \\ 
reservations early and when I arrived he said we apologize we cannot \\ 
provide you with a double bed so he gave me two separate beds even \\ 
though this is the second time I live in this hotel the service the first \\ 
time was better due to less congestion in the hotel\end{tabular} & 0      & 1                                                           \\ \hline
6 & \begin{tabular}[c]{@{}l@{}}
\<مثال واضح لموظفين رايعين في مكان غير مناسب>\\ 
\<طاقم عمل ممتاز جدا الاصوات حول الفندق وداخل الفندق>\\
\<سيه جدا لا يوجد فيش كهربا في مكان مناسب>\\
\<لاستخدام الحاسب الشخصي انترنت سيه جدا>\\  
A clear example of fantastic service in unsuitable location the work crew\\  
very excellent noises around the hotel and in the hotel very bad unavailability\\  
of power plug in a suitable place to use personal computer internet very bad\end{tabular}                                                                                                                                                                                                                                                                                                                                                      & 1      & 0                                                           \\ \hline
\end{tabular}

\end{table}

\begin{table}[]

\caption{Examples of correct classifications of the CNN model (binary classification).} \label{tableMs2}
\begin{tabular}{|l|l|l|l|}
\hline
\# & Review                                                                                                                                                                                                                                                                                                                                                                                                                                                                                                                                                                                                                                                                                                                                                                                                                                      & Rating & \begin{tabular}[c]{@{}l@{}}Model \\ Prediction\end{tabular} \\ \hline
1 & \begin{tabular}[c]{@{}l@{}}
\<ضعيف الغرفه مزعجه السرير غير نظيف>\\ 
Weak the room annoying the bed not clean\end{tabular}                                                                                                                                                                                                                                                                                                                                                                                                                                                                                                                                                                                                                                                                           & 0      & 0                                                          
\\ \hline
2 & \begin{tabular}[c]{@{}l@{}}
\<مخيب للامل النظافه الخدمه الطلبات الفطور ملغي الانترنيت>\\  
\<معدوم في الغرف التلفزيون خربان ما نسمع الادان للصلاه>\\  
\<لا ارغب هذا الفندق في الرحلات القادمه وشكرا>\\
Disappointing cleanliness service requests breakfast\\
canceled internet non-existent in rooms TV is non-functional\\ 
we do not hear call to prayer I do not want to stay at this\\
hotel in my next trips and thank you
\end{tabular}                                                                                                                                                                                                                            & 0      & 0
 \\ \hline
3 & \begin{tabular}[c]{@{}l@{}}
\<رايع السكن مقارنه بالسعر ممتاز المكان حق نوم يعني تنام>\\  
\<مرتاح نظافه وهدو مافي اي وساخه حتي يجيبون لك غساله لو>\\  
\<مطول المكان تحفه مقارنه بالسعر>\\
Fantastic the residence comparing price excellent\\
place for sleeping meaning you sleep comfortable\\ 
quietness and cleanliness no dirt they even bring you a washing\\ machine if staying long place is masterpiece comparing price
\end{tabular}                                                                                                                                                                                                                            & 1      & 1 \\ \hline
4 & \begin{tabular}[c]{@{}l@{}}
\<جميل وهادي وقريب من الخدمات وبه مسبح روعه وجيم واي فاي>\\ 
\<كل شي جميل تقريبا نظافه وهدو ومسبح رايع ماعدا اشيا صفيره>\\
\<المماطله في توفير بعض المستلزمات>\\ 
Beautiful and quiet and near services and has fantastic swimming pool\\ 
and gym Wi-Fi everything beautiful approximately cleanliness and \\ 
quietness and fantastic swimming pool except small things \\
procrastination in providing some amenities\\ 
\end{tabular}                                                                                                                                                                                                                                                                                                                                                                                                                         & 1      & 1                                                           \\ \hline
\end{tabular}

\end{table}

Although the DL classifiers performed exceptionally well it is prudent to find the misclassifications of the model to establish a better understanding of the limitations of the model used and what could further be done to improve the classification. In Table \ref{tableMs}, six misclassification examples of the CNN model are shown, including the review sentences in Arabic and their literal translation in English, along with their true rating and the prediction of the DL models. Keep in mind that these sentences are how they appear after the cleaning process, right before tokenization and padding.\par
The first case presents the important issue of mislabeling by the initial reviewer, where the sentence contains no positive aspects, however, the sentiment is labeled as positive. These errors can only be prevented by vetting the dataset used to remove all mislabeled data, which is not an easy process.\par
The second case presents a positively labeled sentiment that focuses on a single negative aspect of the review, in this instance, the Wi-Fi. Even though the user praises the hotel in the first part of the review, it is only a small part, around 20\% of the total review. The reviewer’s focus on the negative aspect is why we believe the model misclassified this review.\par
The third case presents a review that highlights the shortcomings of the model used, as its prediction was based on the only positive feature in the sentence being the first word, which was misspelled.\par
The fourth case is the opposite of the second case, wherein the reviewer highlights the positive aspects for the majority of the sentence and refers to the negative aspect right at the end, which is the sentiment-deciding factor for the reviewer. The model misclassified this review as a positive review when in fact it is negative.\par 
The fifth case introduces a detailed review that highlights the positive feature (breakfast), with the mention of the word \<(ممتازين)>, meaning excellent. It however fails to emphasize the negative aspects, i.e., no mention of the words (bad, poor, negative). Subsequently, the model misclassified the sentiment of the sentence and confused it with a positive sentiment.\par
The sixth case combined both positive and negative aspects, where the reviewer labeled this review as positive the model classified it as negative.\par
In Table \ref{tableMs2}, four correct classification examples of the CNN model are shown, including the review sentences in Arabic and their literal translation in English, along with their true rating and the prediction of the DL models. We can see that these sentences clearly state the positive or negative aspects and emphasize the sentiment with words such as \<(رايع)> which is the normalized version of \<(رائع)>, meaning fantastic, and \<(ضعيف)> meaning weak.

\section{Conclusion}
In this study, a thorough comparison was conducted between deep and SL techniques for Arabic SA. This comparison is the first of its kind, as previous works only consider a fraction of the techniques proposed in this work. The three main archetypes of DL (CNN, LSTM and GRU), as well as all their possible hybrids, were evaluated against eight of the most-widely used SL classifiers. Furthermore, some state-of-the-art techniques were included such as the TRANS architecture as well as the araBERT pre-trained model. The two datasets used are some of the largest available: the HARD and BRAD datasets. DL techniques outperformed SL techniques in all cases except in 5-label classification, where RF performed best. Furthermore, both techniques performed best when trained and tested using the HARD dataset as compared to the BRAD dataset, even though BRAD was 33\% larger. The reasoning behind this is attributed to the significantly large vocabulary space in BRAD, which was more than twice that of HARD. The best-performing SL techniques were the ensemble learning methods, Random Forest, Decision Tree, and AdaBoost, whereas all the DL techniques performed somewhat similarly.\par
These results show that using DL techniques is more favorable to SL methods for SA of Arabic reviews. This clashes with what was proposed in \cite{al2018deep1}, where it was found that SVM outperformed CNN and RNN in terms of binary sentiment classification. The performance of DL methods improved in our experiments due to the use of larger datasets for training and testing the DL models. It is observed that there is a direct relationship between the performance of DL techniques and the size of the dataset used. To verify this claim, we plotted the learning curve as a function of dataset size. While the increase in accuracy of the DL models diminishes as the dataset increases, it still has a significant effect  on the mean accuracy obtained from multiple runs.\par
Thus we can conclude that for large Arabic datasets (containing tens of thousands of sentences), the use of DL techniques for SA is optimal, whereas for small datasets (less than 10,000 sentences), using a SL technique would be preferred. It must also be highlighted that the TRANS architecture benefits greatly from the use of more intricate embedding layers, particularly pre-trained models such as araBERT. With these advanced embeddings, the TRANS model is able to outperform all other deep and SL techniques for all sizes of datasets except for those that are very small (less than 3,000 sentences).\par
We believe that further treatment of Arabic dialects could enhance the performance of DL techniques for Arabic SA applications. While the performance achieved by the DL classifiers was exemplary, further work can be done to increase the accuracy from the mid-90s to the high-90s.\par

\bibliographystyle{ACM-Reference-Format}
\bibliography{mybibfile}

\section{Appendix}\label{appen}
After various iterations of trial and error, the following configurations consistently performed best and thus were used as the basis of this study.

\begin{table}[htbp]

\caption{TRANS} \label{Apen11}
\begin{tabular}{|l|l|}
\hline
Layer (type) & Output Shape \\ \hline
Embedding & (None, 100, 64) \\ \hline
Transformer & (None, 100, 64) \\ \hline
Global Max Pooling & (None, 64) \\ \hline
Dropout 1 & (None, 64) \\ \hline
Dense 1 & (None, 32) \\ \hline
Dropout 2 & (None, 32) \\ \hline
Dense 2 & (None, 1) \\ \hline
\end{tabular}

\end{table}
\begin{table}[htbp]

\caption{CNN} \label{Apen1}
\begin{tabular}{|l|l|}
\hline
Layer (type)          & Output Shape     \\ \hline
Embedding             & (None, 100, 64)  \\ \hline
Convolutional 1 (1-D) & (None, 100, 300) \\ \hline
Convolutional 2 (1-D) & (None, 100, 64)  \\ \hline
Convolutional 3 (1-D) & (None, 100, 32)  \\ \hline
Global Max Pooling    & (None, 32)       \\ \hline
Dropout               & (None, 32)       \\ \hline
Dense 1               & (None, 32)       \\ \hline
Dense 2               & (None, 1)        \\ \hline
\end{tabular}

\end{table}
\begin{table}[]

\caption{LSTM} \label{Apen2}
\begin{tabular}{|l|l|}
\hline
Layer (type)       & Output Shape    \\ \hline
Embedding          & (None, 100, 64) \\ \hline
Bidirectional LSTM & (None, 64)      \\ \hline
Dense 1            & (None, 32)      \\ \hline
Dense 2            & (None, 1)       \\ \hline
\end{tabular}

\end{table}
\begin{table}[]

\caption{GRU} \label{Apen3}
\begin{tabular}{|l|l|}
\hline
Layer (type) & Output Shape \\ \hline
Embedding & (None, 100, 64) \\ \hline
GRU & (None, 100, 64) \\ \hline
Global Max Pooling & (None, 64) \\ \hline
Dense 1 & (None, 32) \\ \hline
Dense 2 & (None, 1) \\ \hline
\end{tabular}

\end{table}
\begin{table}[]

\caption{CNN+LSTM} \label{Apen4}
\begin{tabular}{|l|l|}
\hline
Layer (type)          & Output Shape    \\ \hline
Embedding             & (None, 100, 64) \\ \hline
Convolutional 1 (1-D) & (None, 100, 64) \\ \hline
Convolutional 2 (1-D) & (None, 100, 32) \\ \hline
Bidirectional LSTM    & (None, 32)      \\ \hline
Dense                 & (None, 1)       \\ \hline
\end{tabular}

\end{table}
\begin{table}[]

\caption{LSTM+CNN} \label{Apen5}
\begin{tabular}{|l|l|}
\hline
Layer (type)          & Output Shape    \\ \hline
Embedding             & (None, 100, 64) \\ \hline
Bidirectional LSTM    & (None, 100, 64) \\ \hline
Convolutional 1 (1-D) & (None, 100, 64) \\ \hline
Convolutional 2 (1-D) & (None, 100, 32) \\ \hline
Global Max Pooling    & (None, 32)      \\ \hline
Dropout               & (None, 32)      \\ \hline
Dense                 & (None, 1)       \\ \hline
\end{tabular}

\end{table}
\begin{table}[]

\caption{GRU+CNN} \label{Apen6}
\begin{tabular}{|l|l|}
\hline
Layer (type)          & Output Shape    \\ \hline
Embedding             & (None, 100, 64) \\ \hline
GRU                   & (None, 100, 64) \\ \hline
Convolutional 1 (1-D) & (None, 100, 64) \\ \hline
Convolutional 2 (1-D) & (None, 100, 32) \\ \hline
Global Max Pooling    & (None, 32)      \\ \hline
Dropout               & (None, 32)      \\ \hline
Dense                 & (None, 1)       \\ \hline
\end{tabular}

\end{table}
\begin{table}[]

\caption{CNN+GRU} \label{Apen7}
\begin{tabular}{|l|l|}
\hline
Layer (type) & Output Shape \\ \hline
Embedding & (None, 100, 64) \\ \hline
Convolutional 1 (1-D) & (None, 100, 64) \\ \hline
Convolutional 2 (1-D) & (None, 100, 32) \\ \hline
GRU & (None, 100, 32) \\ \hline
Global Max Pooling & (None, 32) \\ \hline
Dense & (None, 1) \\ \hline
\end{tabular}

\end{table}
\begin{table}[]

\caption{LSTM+GRU} \label{Apen8}
\begin{tabular}{|l|l|}
\hline
Layer (type)       & Output Shape    \\ \hline
Embedding          & (None, 100, 64) \\ \hline
Bidirectional LSTM & (None, 100, 64) \\ \hline
GRU                & (None,  32)     \\ \hline
Dense 1            & (None, 1)       \\ \hline
\end{tabular}

\end{table}
\begin{table}[]

\caption{GRU+LSTM} \label{Apen9}
\begin{tabular}{|l|l|}
\hline
Layer (type)       & Output Shape    \\ \hline
Embedding          & (None, 100, 64) \\ \hline
GRU                & (None, 100, 64) \\ \hline
Bidirectional LSTM & (None,  32)     \\ \hline
Dense 1            & (None, 1)       \\ \hline
\end{tabular}

\end{table}
\begin{table}[]

\caption{Shallow Learning Classifiers} \label{Apen10}
\begin{tabular}{|c|l|}
\hline
Name & \multicolumn{1}{c|}{Parameters} \\ \hline
\begin{tabular}[c]{@{}c@{}}Decision\\ Tree\end{tabular} & \begin{tabular}[c]{@{}l@{}}ccp\_alpha=0.0, class\_weight=None, criterion='gini', max\_depth=None, \\ max\_features=None, max\_leaf\_nodes=None, min\_impurity\_decrease=0.0, \\ min\_impurity\_split=None, min\_samples\_leaf=1, min\_samples\_split=2, \\ min\_weight\_fraction\_leaf=0.0, presort='deprecated', random\_state=None, \\ splitter='best'\end{tabular} \\ \hline
\begin{tabular}[c]{@{}c@{}}Random\\ Forest\end{tabular} & \begin{tabular}[c]{@{}l@{}}bootstrap=True, ccp\_alpha=0.0, class\_weight=None, criterion='gini', \\ max\_depth=None, max\_features='auto', max\_leaf\_nodes=None, \\ max\_samples=None, min\_impurity\_decrease=0.0, \\ min\_impurity\_split=None, min\_samples\_leaf=1, min\_samples\_split=2, \\ min\_weight\_fraction\_leaf=0.0, n\_estimators=100, n\_jobs=None, \\ oob\_score=False, random\_state=None, verbose=0, warm\_start=False\end{tabular} \\ \hline
\begin{tabular}[c]{@{}c@{}}SVM\\ (RBF)\end{tabular} & \begin{tabular}[c]{@{}l@{}}C=1.0, break\_ties=False, cache\_size=200, class\_weight=None, \\ coef0=0.0, decision\_function\_shape='ovr', degree=3, \\ gamma='scale', kernel='rbf', max\_iter=-1, probability=False, \\ random\_state=None, shrinking=True, tol=0.001, verbose=False\end{tabular} \\ \hline
\begin{tabular}[c]{@{}c@{}}SVM\\ (Linear)\end{tabular} & \begin{tabular}[c]{@{}l@{}}C=1.0, class\_weight=None, dual=True, fit\_intercept=True, \\ intercept\_scaling=1, loss='squared\_hinge', max\_iter=1000, \\ multi\_class='ovr', penalty='l2', random\_state=None, tol=0.0001, \\ verbose=0\end{tabular} \\ \hline
KNN & \begin{tabular}[c]{@{}l@{}}algorithm='auto', leaf\_size=30, metric='minkowski', \\ metric\_params=None, n\_jobs=None, n\_neighbors=5, \\ p=2, weights='uniform'\end{tabular} \\ \hline
MLP & \begin{tabular}[c]{@{}l@{}}activation='relu', alpha=0.0001, batch\_size='auto', beta\_1=0.9, \\ beta\_2=0.999, early\_stopping=False, epsilon=1e-08, \\ hidden\_layer\_sizes=(100,), learning\_rate='constant', \\ learning\_rate\_init=0.001, max\_fun=15000, max\_iter=200, \\ momentum=0.9, n\_iter\_no\_change=10, nesterovs\_momentum=True, \\ power\_t=0.5, random\_state=None, shuffle=True, solver='adam', \\ tol=0.0001, validation\_fraction=0.1, verbose=False, \\ warm\_start=False\end{tabular} \\ \hline
\begin{tabular}[c]{@{}c@{}}Gaussian\\ NB\end{tabular} & priors=None, var\_smoothing=1e-09 \\ \hline
AdaBoost & \begin{tabular}[c]{@{}l@{}}algorithm='SAMME.R', base\_estimator=None, \\ learning\_rate=1.0, n\_estimators=50, random\_state=None\end{tabular} \\ \hline
\end{tabular}

\end{table}

\end{document}